\documentclass[letterpaper,10pt,conference]{ieeeconf}
\IEEEoverridecommandlockouts
\usepackage{times}
\usepackage{cite}
\usepackage[utf8]{inputenc} 
\usepackage[T1]{fontenc}    
\usepackage[draft]{hyperref}       
\usepackage{url}            
\usepackage{booktabs}       
\usepackage{amsfonts}       
\usepackage{nicefrac}       
\usepackage{microtype}      
\PassOptionsToPackage{table}{xcolor}
\usepackage{xcolor}         
\usepackage{colortbl}
\makeatletter
\newcommand{\oursrowbackground}{\noalign{\begingroup\color{gray!10}%
  \hrule width\linewidth height\dimexpr\ht\@arstrutbox+\dp\@arstrutbox\relax
  \kern-\dimexpr\ht\@arstrutbox+\dp\@arstrutbox\relax\endgroup}}
\makeatother

\usepackage{amsmath}
\usepackage{amssymb}
\usepackage{mathtools}
\usepackage{graphicx}
\usepackage{multirow}
\usepackage{tikz}
\usetikzlibrary{positioning, arrows.meta, calc, fit, decorations.pathreplacing, shapes.geometric, shadows.blur, backgrounds}
\definecolor{figInk}{HTML}{273449}
\definecolor{figBlue}{HTML}{3278CE}
\definecolor{figTeal}{HTML}{218C88}
\definecolor{figViolet}{HTML}{8665BB}
\definecolor{figRose}{HTML}{C86D81}
\definecolor{figAmber}{HTML}{C58A3C}
\tikzset{
  fig shadow/.style={blur shadow={shadow xshift=0.5pt, shadow yshift=-1.1pt,
    shadow blur radius=2pt, shadow opacity=16, shadow blur steps=8}},
  fig surface/.style={rounded corners=4pt, line width=0.45pt,
    text=figInk, fig shadow},
  fig panel/.style={fig surface, draw=figInk!12,
    top color=white, bottom color=figBlue!3},
}
\newcommand{\figpanel}[2]{%
  \begin{scope}[on background layer]
    \path[fig panel] #1 rectangle #2;
  \end{scope}%
}

\let\labelindent\relax
\usepackage{enumitem}
\usepackage{xspace}
\usepackage{float}
\usepackage{wrapfig}
\usepackage{needspace}

\usepackage[capitalize]{cleveref}
\crefname{section}{Sec.}{Secs.}
\crefname{table}{Table}{Tables}
\crefname{figure}{Fig.}{Figs.}
\crefname{appendix}{Suppl.}{Suppls.}

\creflabelformat{figure}{#2\textcolor{red}{#1}#3}
\creflabelformat{table}{#2\textcolor{red}{#1}#3}
\creflabelformat{section}{#2\textcolor{red}{#1}#3}
\creflabelformat{subsection}{#2\textcolor{red}{#1}#3}
\creflabelformat{subsubsection}{#2\textcolor{red}{#1}#3}
\creflabelformat{appendix}{#2\textcolor{red}{#1}#3}

\makeatletter
\renewcommand{\@IEEEfigurecaptionsepspace}{\vskip 3pt\relax}
\makeatother

\title{\LARGE\bf Ada3Drift: Adaptive Training-Time Drifting for One-Step 3D Visuomotor Robotic Manipulation}
\author{Chongyang Xu$^1$, Yixian Zou$^2$, Tianyu Yang$^2$, Fanman Meng$^2$, Ziliang Feng$^1$, Li Lu$^{1,*}$, and Shuaicheng Liu$^{2,*}$%
\thanks{$^1$College of Computer Science, Sichuan University.}%
\thanks{$^2$University of Electronic Science and Technology of China.}%
\thanks{$^*$Corresponding authors.}%
}
\begin{document}
\maketitle
\thispagestyle{empty}
\pagestyle{empty}
\begin{abstract}
Diffusion-based visuomotor policies model complex action distributions through iterative denoising, but repeated inference adds latency to robotic control. One-step generators reduce this cost, motivating training objectives that retain useful action structure with few demonstrations. We present Ada3Drift, a point-cloud-conditioned policy that builds on Drifting Models to perform distribution refinement during training and generate action chunks in one forward pass. Our central design is a regression-to-drifting curriculum: paired action regression first emphasizes observation--action correspondence, while a sigmoid schedule progressively increases a batch-level action-distribution regularizer. The drifting term combines attraction to demonstrated actions and repulsion among generated samples using inherited multi-temperature aggregation. A timestep-free generator preserves single-step inference throughout. On Adroit, Meta-World, RoboTwin, and five real-world tasks, Ada3Drift achieves the highest reported average success rates among the evaluated baselines with 1 NFE, compared with 10 NFE for the diffusion baselines. Controlled ablations favor sigmoid scheduling at matched demonstration budgets. We will release our code and pretrained model weights.
\end{abstract}

\section{Introduction}
\label{sec:intro}

Learning visuomotor policies from expert demonstrations has achieved remarkable progress~\cite{brohan2023rt1roboticstransformerrealworld, zitkovich2023rt2, jang2022bcz, florence2022implicit, bharadhwaj2024track2act, 10.1007/978-3-031-72652-1_21}. A central challenge is handling the \textit{multimodal} nature of human demonstrations: for a given observation, multiple valid action strategies may exist. Diffusion Policy~\cite{chi2023diffusion} and its 3D extension DP3~\cite{ze20243d} addressed this challenge through iterative denoising, which progressively refines noisy predictions and can represent distinct action modes. This capability has driven a wave of diffusion-based robotic systems~\cite{ma2024hierarchical, octo2024, rdt2024, black2024pi0, Vosylius2024InstantPI}, yet sampling commonly uses 10--100 function evaluations (NFE) per action chunk, increasing inference latency in \mbox{real-time robotic manipulation.}

Recent approaches based on flow matching~\cite{zhang2025flowpolicy, mp1, hu2024adaflow}, consistency distillation~\cite{prasad2024consistency, lu2024manicm}, and diffusion distillation~\cite{wang2024onestep, Jia2024ScoreAD} reduce generation to one or a few forward passes. A single coarse integration step of a standard flow can introduce approximation error, whereas specialized objectives such as MeanFlow~\cite{geng2025meanflowsonestepgenerative} learn a field suited to one-step generation. Single-step inference therefore does not inherently imply mode averaging. Nevertheless, preserving useful action structure with limited demonstrations remains important: directly regressing ambiguous actions can average distinct strategies, and the average of two collision-free trajectories need not itself be feasible (\cref{fig:comparison}). This motivates investigating training objectives that combine observation--action correspondence with distribution-level information.

Robotic systems exhibit a useful \textbf{compute budget asymmetry}: optimization is performed offline, whereas policy inference must fit a real-time execution loop. This motivates placing distribution refinement in the training objective while retaining a single-step generator. Our focus is how to balance two sources of supervision under limited data. Paired regression provides a direct signal for predicting the demonstrated action from its observation. Distribution-level interactions additionally compare generated and demonstrated action populations. Their relative influence need not remain constant throughout training. We investigate a preset curriculum that changes this balance over epochs, rather than adapting online to task identity or model uncertainty.
\begin{figure}[t]
\centering
\resizebox{\columnwidth}{!}{%
\begin{tikzpicture}[>=Stealth, text=figInk]


\foreach \panelx in {0,3,6} {
  \begin{scope}[xshift=\panelx cm]
    \figpanel{(-1.23,-1.22)}{(1.35,2.62)}
  \end{scope}
}

\newcommand{\drawTopGauss}[1]{%
  \shade[top color=cyan!20, bottom color=cyan!2] plot[domain=-1.1:1.1, samples=50, smooth]
    ({#1+\x}, {2.0 + 0.45*exp(-\x*\x/0.25)}) -- ({#1+1.1},2.0) -- ({#1-1.1},2.0) -- cycle;
  \draw[cyan!60!blue, semithick] plot[domain=-1.1:1.1, samples=50, smooth]
    ({#1+\x}, {2.0 + 0.45*exp(-\x*\x/0.25)});
}

\newcommand{\drawBotBimodal}[1]{%
  \shade[top color=orange!2, bottom color=orange!20] plot[domain=-1.1:1.1, samples=60, smooth]
    ({#1+\x}, {-0.45*(exp(-(\x-0.5)^2/0.05) + exp(-(\x+0.5)^2/0.05))})
    -- ({#1+1.1},0) -- ({#1-1.1},0) -- cycle;
  \draw[orange!70!red, semithick] plot[domain=-1.1:1.1, samples=60, smooth]
    ({#1+\x}, {-0.45*(exp(-(\x-0.5)^2/0.05) + exp(-(\x+0.5)^2/0.05))});
}

\begin{scope}[xshift=0cm]
  \drawTopGauss{0}
  \drawBotBimodal{0}

  \foreach \sx/\op in {-0.6/0.25, -0.3/0.35, 0/0.45, 0.3/0.35, 0.6/0.25} {
    \draw[->, gray!50, semithick, opacity=\op] (\sx, 1.9) -- (0, 0.1);
  }
  \node[red!60, font=\normalsize\bfseries] at (0, 0.1) {$\times$};

  \node[font=\footnotesize\bfseries, anchor=north] at (0, -0.75) {(a) Paired Regression};
  \node[font=\scriptsize, red!50, anchor=west] at (0.3, 0.25) {mode avg.};
\end{scope}

\begin{scope}[xshift=3.0cm]
  \drawTopGauss{0}
  \drawBotBimodal{0}

  \draw[->, gray!50, semithick, opacity=0.25] (-0.6, 1.9) to[out=-80, in=115] (-0.15, 0.1);
  \draw[->, gray!50, semithick, opacity=0.35] (-0.3, 1.9) to[out=-83, in=108] (-0.08, 0.1);
  \draw[->, gray!50, semithick, opacity=0.45] (0, 1.9) to[out=-87, in=93] (0, 0.1);
  \draw[->, gray!50, semithick, opacity=0.35] (0.3, 1.9) to[out=-97, in=72] (0.08, 0.1);
  \draw[->, gray!50, semithick, opacity=0.25] (0.6, 1.9) to[out=-100, in=65] (0.15, 0.1);

  \node[orange!70!red, font=\normalsize\bfseries] at (0, 0.1) {$\triangle$};

  \node[font=\footnotesize\bfseries, anchor=north] at (0, -0.75) {(b) Coarse Flow Step};
  \node[font=\scriptsize, orange!60!red, anchor=west] at (0.3, 0.25) {biased};
\end{scope}

\begin{scope}[xshift=6.0cm]
  \drawTopGauss{0}
  \drawBotBimodal{0}

  \draw[gray!35, dashed, semithick] (-0.6, 1.9) -- (-0.35, 1.05);
  \draw[->, black!70, semithick] (-0.35, 1.05) to[out=-100, in=80] (-0.5, 0.1);
  \draw[gray!35, dashed, semithick] (-0.3, 1.9) -- (-0.15, 1.05);
  \draw[->, black!70, semithick] (-0.15, 1.05) to[out=-105, in=75] (-0.45, 0.1);
  \draw[gray!35, dashed, semithick] (0, 1.9) -- (0, 1.05);
  \draw[->, black!70, semithick] (0, 1.05) to[out=-110, in=70] (-0.42, 0.1);

  \draw[gray!35, dashed, semithick] (0.3, 1.9) -- (0.15, 1.05);
  \draw[->, black!70, semithick] (0.15, 1.05) to[out=-75, in=105] (0.45, 0.1);
  \draw[gray!35, dashed, semithick] (0.6, 1.9) -- (0.35, 1.05);
  \draw[->, black!70, semithick] (0.35, 1.05) to[out=-70, in=110] (0.5, 0.1);

  \draw[figViolet!55, fill=figViolet!7, fill opacity=0.3, semithick, dashed, rounded corners=3pt] (-0.85, 1.28) rectangle (0.85, 0.82);
  \node[font=\scriptsize, black!60] at (0, 1.4) {$V(\hat{\mathbf{x}})$};

  \node[green!60!black, font=\small\bfseries] at (-0.75, 0.35) {\checkmark};
  \node[green!60!black, font=\small\bfseries] at (0.75, 0.35) {\checkmark};

  \node[font=\footnotesize\bfseries, black!80, anchor=north] at (0, -0.75) {(c) Drift Training};
\end{scope}

\node[font=\scriptsize, cyan!60!blue] at (-1.5, 2.25) {$p_0$};
\node[font=\scriptsize, orange!70!red] at (-1.5, -0.15) {$p_1$};
\draw[->, gray!30, semithick] (-1.5, 2.0) -- (-1.5, 0.1);
\node[font=\scriptsize, gray!40, rotate=90, anchor=south] at (-1.7, 1.0) {generation};

\end{tikzpicture}%
}
\caption{\textbf{Motivation for training-time distribution regularization.}
(a)~Paired regression can average ambiguous actions.
(b)~A coarse step of a standard flow can incur integration error.
(c)~Attraction and repulsion provide distribution-level training signals for a one-step generator. Paths are conceptual, not measured trajectories. These conceptual failure modes do not necessarily apply to methods explicitly trained for one-step generation.}
\label{fig:comparison}
\end{figure}

\begin{figure*}[t]
\centering
\resizebox{0.9\textwidth}{!}{%
\begin{tikzpicture}[>=Stealth, text=figInk]


\begin{scope}[xshift=0cm]
  \node[font=\footnotesize\bfseries] at (3.5, 3.6) {(a) Drifting Field $V(\tilde{\mathbf{x}})$ with Multi-$\tau$ Aggregation};

  \begin{scope}[on background layer]
    \shade[inner color=cyan!13, outer color=white] (1.5,2.7) ellipse (1.5 and 0.85);
    \shade[inner color=cyan!13, outer color=white] (5.5,0.85) ellipse (1.4 and 0.75);
  \end{scope}
  \draw[cyan!12, thick, dashed] (1.5, 2.7) ellipse (1.5 and 0.85);
  \draw[cyan!28, thick] (1.5, 2.7) ellipse (0.9 and 0.5);
  \draw[cyan!55!blue, thick] (1.5, 2.7) ellipse (0.4 and 0.2);

  \draw[cyan!12, thick, dashed] (5.5, 0.85) ellipse (1.4 and 0.75);
  \draw[cyan!28, thick] (5.5, 0.85) ellipse (0.95 and 0.5);
  \draw[cyan!55!blue, thick] (5.5, 0.85) ellipse (0.55 and 0.27);

  \fill[orange!60] (1.2, 2.9) circle (4pt) coordinate (e1);
  \fill[orange!60] (1.7, 2.55) circle (4pt) coordinate (e2);
  \fill[orange!60] (1.3, 2.45) circle (4pt) coordinate (e2b);
  \draw[orange!80!black, semithick] (e1) circle (4pt);
  \draw[orange!80!black, semithick] (e2) circle (4pt);
  \draw[orange!80!black, semithick] (e2b) circle (4pt);

  \fill[orange!60] (5.2, 0.95) circle (4pt) coordinate (e3);
  \fill[orange!60] (5.7, 1.25) circle (4pt) coordinate (e4);
  \fill[orange!60] (5.5, 0.85) circle (4pt) coordinate (e4b);
  \draw[orange!80!black, semithick] (e3) circle (4pt);
  \draw[orange!80!black, semithick] (e4) circle (4pt);
  \draw[orange!80!black, semithick] (e4b) circle (4pt);

  \fill[cyan!70!blue] (2.8, 2.2) circle (3.5pt) coordinate (p1);
  \fill[cyan!70!blue] (3.2, 1.85) circle (3.5pt) coordinate (p2);
  \fill[cyan!70!blue] (3.5, 2.35) circle (3.5pt) coordinate (p3);
  \fill[cyan!70!blue] (3.8, 1.55) circle (3.5pt) coordinate (p4);
  \fill[cyan!70!blue] (3.3, 1.35) circle (3.5pt) coordinate (p5);
  \fill[cyan!70!blue] (4.0, 2.05) circle (3.5pt) coordinate (p6);
  \fill[cyan!70!blue] (3.0, 1.5) circle (3.5pt) coordinate (p7);

  \foreach \dot in {e1,e2,e2b,e3,e4,e4b,p1,p2,p3,p4,p5,p6,p7} {
    \fill[white, opacity=0.65] ([xshift=-0.9pt,yshift=1.1pt]\dot) circle (0.8pt);
  }

  \draw[->, orange!70!red, semithick] (p1) -- ($(e2)!0.15!(p1)$);
  \draw[->, orange!70!red, semithick] (p2) -- ($(e2b)!0.15!(p2)$);
  \draw[->, orange!70!red, semithick] (p3) -- ($(e1)!0.15!(p3)$);
  \draw[->, orange!70!red, semithick] (p4) -- ($(e3)!0.15!(p4)$);
  \draw[->, orange!70!red, semithick] (p5) -- ($(e3)!0.15!(p5)$);
  \draw[->, orange!70!red, semithick] (p6) -- ($(e4)!0.15!(p6)$);
  \draw[->, orange!70!red, semithick] (p7) -- ($(e4b)!0.15!(p7)$);

  \draw[->, cyan!50!blue, semithick, opacity=0.55] (p1) -- ++(-.28, .3);
  \draw[->, cyan!50!blue, semithick, opacity=0.55] (p2) -- ++(.25, -.28);
  \draw[->, cyan!50!blue, semithick, opacity=0.55] (p3) -- ++(-.22, .28);
  \draw[->, cyan!50!blue, semithick, opacity=0.55] (p4) -- ++(.28, -.22);
  \draw[->, cyan!50!blue, semithick, opacity=0.55] (p5) -- ++(-.25, -.22);
  \draw[->, cyan!50!blue, semithick, opacity=0.55] (p6) -- ++(.22, .25);

  \draw[cyan!55!blue, thin] (5.9, 1.1) -- (6.6, 1.7) node[right, font=\scriptsize, cyan!55!blue] {$\tau\!=\!0.02$};
  \draw[cyan!35!black, thin] (6.4, 1.1) -- (6.6, 0.95) node[right, font=\scriptsize, cyan!35!black] {$\tau\!=\!0.05$};
  \draw[cyan!15!black, thin] (6.9, 0.7) -- (7.0, 0.15) node[right, font=\scriptsize, cyan!15!black] {$\tau\!=\!0.2$};

  \fill[orange!60] (0.3, 0.2) circle (3.5pt);
  \draw[orange!80!black, semithick] (0.3, 0.2) circle (3.5pt);
  \node[font=\scriptsize, anchor=west] at (0.55, 0.2) {expert $\tilde{\mathbf{y}}^+$};
  \fill[cyan!70!blue] (2.0, 0.2) circle (3pt);
  \node[font=\scriptsize, anchor=west] at (2.25, 0.2) {pred.\ $\tilde{\mathbf{x}}$};
  \draw[->, orange!70!red, semithick] (3.7, 0.2) -- (4.15, 0.2);
  \node[font=\scriptsize, anchor=west] at (4.25, 0.2) {attract};
  \draw[->, cyan!50!blue, semithick] (5.5, 0.2) -- (5.95, 0.2);
  \node[font=\scriptsize, anchor=west] at (6.05, 0.2) {repel};
\end{scope}

\begin{scope}[xshift=8.6cm, xscale=1.5]
  \node[font=\footnotesize\bfseries] at (1.5, 3.6) {(b) Sigmoid Schedule};

  \draw[->, gray!40, semithick] (0, 0.6) -- (3.2, 0.6);
  \draw[->, gray!40, semithick] (0, 0.6) -- (0, 3.4);
  \node[font=\scriptsize, gray!50] at (3.7, 0.4) {epoch};
  \node[font=\scriptsize, gray!50, rotate=90] at (-0.25, 2.0) {weight};

  \foreach \lw/\mix in {3.6pt/5,2.4pt/9} {
    \draw[cyan!\mix, line width=\lw] plot[domain=0:3, samples=70, smooth]
      ({\x}, {0.6 + 2.4*(\x/3 < 0.7 ? 1/(1+exp(-20*abs(\x/3 - 0.7))) : exp(-20*abs(\x/3 - 0.7))/(1+exp(-20*abs(\x/3 - 0.7))))});
    \draw[orange!\mix, line width=\lw] plot[domain=0:3, samples=70, smooth]
      ({\x}, {0.6 + 2.4*(1-(\x/3 < 0.7 ? 1/(1+exp(-20*abs(\x/3 - 0.7))) : exp(-20*abs(\x/3 - 0.7))/(1+exp(-20*abs(\x/3 - 0.7)))))});
  }
  \draw[cyan!60!blue, thick] plot[domain=0:3, samples=70, smooth]
    ({\x}, {0.6 + 2.4*(\x/3 < 0.7 ? 1/(1+exp(-20*abs(\x/3 - 0.7))) : exp(-20*abs(\x/3 - 0.7))/(1+exp(-20*abs(\x/3 - 0.7))))});

  \draw[orange!70!red, thick] plot[domain=0:3, samples=70, smooth]
    ({\x}, {0.6 + 2.4*(1-(\x/3 < 0.7 ? 1/(1+exp(-20*abs(\x/3 - 0.7))) : exp(-20*abs(\x/3 - 0.7))/(1+exp(-20*abs(\x/3 - 0.7)))))});

  \draw[gray!25, dashed, thin] (2.1, 0.6) -- (2.1, 3.2);
  \node[font=\scriptsize, gray!45] at (2.1, 0.4) {$0.7E$};

  \node[font=\scriptsize, gray!65, anchor=east] at (-0.04,0.6) {$0$};
  \node[font=\scriptsize, gray!65, anchor=east] at (-0.04,3.0) {$1$};
  \node[font=\scriptsize, gray!65, anchor=north] at (3.0,0.57) {$E$};
  \node[font=\scriptsize, cyan!60!blue] at (0.55, 3.2) {$w_{\text{mse}}$};
  \node[font=\scriptsize, orange!70!red] at (2.7, 3.2) {$w_{\text{drift}}$};

  \node[font=\scriptsize, cyan!50!blue] at (1.0, 0.2) {regression};
  \node[font=\scriptsize, orange!55!red] at (2.5, 0.2) {drifting};
\end{scope}

\node[font=\scriptsize, text=figInk] at (2.05,0.76) {
  $V(\tilde{\mathbf{x}}_i)=\textcolor{orange!70!red}{\sum_j W^+_{ij}\tilde{\mathbf{y}}^+_j}
  -\textcolor{cyan!60!blue}{\sum_k W^-_{ik}\tilde{\mathbf{x}}_k}$};
\node[fig panel, font=\scriptsize, inner sep=4pt] at (5.65,2.85) {
  $A_{ij}=\sqrt{A^{\mathrm{row}}_{ij}A^{\mathrm{col}}_{ij}}$};
\node[fig panel, font=\scriptsize, align=center, inner sep=4pt] at (10.0,1.65) {
  $w_{\mathrm{drift}}(e)=\sigma(u)$\\[2pt]
  $u=\frac{e-0.7E}{0.05E}$\\[2pt]
  $w_{\mathrm{mse}}=1-w_{\mathrm{drift}}$};
\end{tikzpicture}%
}
\caption{\textbf{Drifting regularization and the training curriculum.}
(a)~Expert attraction and generated-sample repulsion act across a task minibatch. Contours schematically illustrate kernel scales $\tau\in\{0.02,0.05,0.2\}$ after distance normalization, with effective temperature $\tau\sqrt D$. They do not represent measured conditional modes.
(b)~The preset sigmoid shifts weight from paired MSE to drifting, crossing at $0.7E$.}
\label{fig:drifting}
\end{figure*}

Drifting Models~\cite{deng2026generativemodelingdrifting} already establish training-time distribution evolution for one-step generation and demonstrate robotic control. Building on this foundation, we ask how a point-cloud-conditioned policy should combine paired supervision and drifting when demonstrations are scarce. In our implementation, drifting compares action samples across observations within a task minibatch. It acts as an action-distribution regularizer, while paired MSE preserves the link between each observation and its demonstration. We hypothesize that emphasizing this predictive mapping first provides a useful starting point for later distribution regularization. This motivates a regression-to-drifting curriculum, whose benefit we evaluate against direct drifting, regression alone, and alternative schedules.

We instantiate this approach as \textbf{Ada3Drift} (\textbf{Ada}ptive \textbf{3D} Training-Time \textbf{Drift}ing). A sigmoid schedule progressively increases the drifting weight in a point-cloud-conditioned, timestep-free action generator. Here, adaptive refers to the changing training weights under a fixed schedule. We inherit attraction, repulsion, and multi-temperature aggregation from Drifting Models. Our contribution concerns their combination with paired regression and its controlled evaluation for few-demonstration 3D manipulation.
We summarize \mbox{our contributions as follows:}
\begin{itemize}
    \item We formulate few-demonstration one-step policy learning as a combination of paired observation--action regression and batch-level action-distribution regularization, retaining training-time refinement without iterative inference.
    \item We develop a sigmoid regression-to-drifting curriculum for a point-cloud-conditioned generator and compare it with fixed and alternative schedules that approximately match cumulative drift weights.
    \item Experiments on Adroit, Meta-World, Robo\-Twin, and real-world manipulation show the highest reported average success rates among the evaluated baselines with 1 NFE. Objective and demonstration-budget ablations characterize the curriculum's performance and limitations.
\end{itemize}

\begin{figure*}[t]
\centering
\resizebox{\textwidth}{!}{%
\begin{tikzpicture}[
    >=Stealth,
    node distance=0.4cm,
    block/.style={rectangle, draw, fig surface, minimum height=0.8cm, minimum width=1.6cm, align=center, font=\footnotesize},
    enc/.style={block, draw=figBlue!48, top color=figBlue!3, bottom color=figBlue!18},
    dec/.style={block, draw=figTeal!48, top color=figTeal!3, bottom color=figTeal!18},
    bot/.style={block, draw=figAmber!48, top color=figAmber!3, bottom color=figAmber!18},
    io/.style={block, top color=white, bottom color=figInk!5, draw=figInk!25, dashed},
    cond/.style={block, draw=figViolet!48, top color=figViolet!3, bottom color=figViolet!18},
    loss/.style={block, draw=figRose!48, top color=figRose!3, bottom color=figRose!18},
    arrow/.style={->, thick, figInk!65},
    skip/.style={->, dashed, gray!50},
    lbl/.style={font=\scriptsize\bfseries},
]


\node[io] (input) {$\mathbf{z} \!\sim\! \mathcal{N}(\mathbf{0},\mathbf{I})$\\[-2pt]{\scriptsize $D_a \!\times\! 16$}};

\node[enc, right=0.5cm of input] (conv_in) {Conv1D\\[-2pt]{\scriptsize $D_a \!\to\! 512$}};
\node[enc, right=0.5cm of conv_in] (enc1) {ResBlock$\times$2\\[-2pt]{\scriptsize $1024\!\times\!16$}};
\node[enc, right=0.5cm of enc1] (enc2) {Downsample\\[-2pt]{\scriptsize $T:16\!\to\!8$}};
\node[enc, right=0.5cm of enc2] (enc3) {ResBlock$\times$2\\[-2pt]{\scriptsize $2048\!\times\!8$}};
\node[bot, right=0.5cm of enc3] (bottleneck) {Bottleneck\\[-2pt]{\scriptsize $2048\!\times\!8$}};

\draw[arrow] (input) -- (conv_in);
\draw[arrow] (conv_in) -- (enc1);
\draw[arrow] (enc1) -- (enc2);
\draw[arrow] (enc2) -- (enc3);
\draw[arrow] (enc3) -- (bottleneck);

\node[font=\tiny, gray, below=0.05cm of enc1] {$T\!=\!16$};
\node[font=\tiny, gray, below=0.05cm of enc2] {$T\!=\!8$};
\node[font=\tiny, gray, below=0.05cm of enc3] {$T\!=\!8$};

\node[font=\scriptsize\itshape, blue!60, above=0.15cm of enc2] {Encoder};
\node[font=\scriptsize\bfseries, red!60!black, above=0.15cm of bottleneck] {No timestep};

\node[dec, below=0.95cm of enc3] (dec3) {ResBlock$\times$2\\[-2pt]{\scriptsize $\to$1024}};
\node[dec, below=0.95cm of enc2] (dec2) {Upsample\\[-2pt]{\scriptsize $T:8\!\to\!16$}};
\node[dec, below=0.95cm of enc1] (dec1) {ResBlock$\times$2\\[-2pt]{\scriptsize $\to$512}};
\node[dec, below=0.95cm of conv_in] (conv_out) {Conv1D\\[-2pt]{\scriptsize $512 \!\to\! D_a$}};

\node[io, below=0.95cm of input] (output) {$\hat{\mathbf{x}}$\\[-2pt]{\scriptsize $D_a \!\times\! 16$}};

\draw[arrow] (bottleneck.south) -- (dec3.north);

\draw[arrow] (dec3) -- (dec2);
\draw[arrow] (dec2) -- (dec1);
\draw[arrow] (dec1) -- (conv_out);
\draw[arrow] (conv_out) -- (output);

\node[font=\scriptsize\itshape, teal!60, below=0.15cm of dec2] {Decoder};

\draw[skip] (enc3.south) -- node[right, font=\tiny, gray] {cat} (dec3.north);
\draw[skip] (enc1.south) -- node[right, font=\tiny, gray] {cat} (dec1.north);

\node[font=\scriptsize, text=figBlue, align=center] at ($(input)!0.5!(output)$)
  {$\hat{\mathbf{x}}=f_\theta(\mathbf{z},\mathbf{g})$};

\node[io, below=0.65cm of output] (gt) {Expert $\mathbf{y}^+$};

\node[loss, right=0.4cm of gt] (drift_field) {Drifting Field $V(\tilde{\mathbf{x}})$\\[-2pt]{\scriptsize expert attraction / generated repulsion}};

\node[loss, right=0.4cm of drift_field] (multitemp) {Multi-$\tau$ Aggregation\\[-2pt]{\scriptsize $\tau\!=\!\{.02,.05,.2\}$}};

\node[loss, right=0.4cm of multitemp] (total_loss) {$\mathcal{L} \!=\! w_d \mathcal{L}_{\text{drift}} \!+\! w_m \mathcal{L}_{\text{mse}}$\\[-2pt]{\scriptsize sigmoid schedule}};

\draw[arrow, rounded corners=3pt] (output.south) -- ++(0,-0.25) -| (drift_field.north);
\draw[arrow] (gt.east) -- (drift_field.west);
\draw[arrow] (drift_field) -- (multitemp);
\draw[arrow] (multitemp) -- (total_loss);

\draw[->, thick, red!40, dashed] (total_loss.north) -- ++(0, 0.35) node[right, font=\scriptsize, red!50!black] {$\nabla_\theta$};

\node[font=\scriptsize\itshape, red!50!black] at ([yshift=-0.35cm]drift_field.south) {Training only ($\mathrm{sg}$: stop-gradient)};

\node[font=\scriptsize, text=figRose!85!black, anchor=north]
  at ([xshift=0.5cm,yshift=-0.52cm]drift_field.south)
  {$\tilde{\mathbf{x}}_{\mathrm{target}}=\mathrm{sg}\!\left(\tilde{\mathbf{x}}+V_{\mathrm{total}}\right)$};

\node[font=\tiny, red!40!black, anchor=north] at ([yshift=-0.15cm]total_loss.south) {$w_d(e) \!=\! \sigma\!\left(\!\frac{e - 0.7E}{0.05E}\!\right)$};

\coordinate (divider_top) at ([xshift=0.75cm]bottleneck.north east);
\coordinate (divider_bot) at (divider_top |- total_loss.south);
\draw[gray!40, dashed, thick] ([yshift=0.6cm]divider_top) -- ([yshift=-0.5cm]divider_bot);


\node[cond, right=1.5cm of bottleneck, yshift=0.1cm] (obs_enc) {Obs Encoder\\[-2pt]{\scriptsize PointNet + MLP}};
\node[io, minimum width=1.2cm, right=0.35cm of obs_enc] (obs_in) {$\mathbf{P}_t,\; \mathbf{s}_t$};
\node[cond, minimum height=0.95cm, inner ysep=5pt, below=0.2cm of obs_enc] (film) {FiLM $(\gamma, \beta)$\\[3pt]{\scriptsize $\mathbf{g} \!\in\! \mathbb{R}^{256}$}};

\draw[arrow] (obs_in) -- (obs_enc);
\draw[arrow] (obs_enc) -- (film);
\draw[->, violet!50, thick, dashed] (film.west) -- ++(-0.8, 0) node[above, font=\tiny, violet!50] {to all ResBlocks};

\node[rectangle, fig surface, top color=white, bottom color=figViolet!8, draw=figViolet!30,
      minimum width=2.8cm, minimum height=2.4cm, align=center, font=\scriptsize,
      below=0.25cm of film] (detail) {
\begin{tabular}{c}
\textbf{Residual Block} \\[3pt]
Conv1D ($k\!=\!5$) \\
GroupNorm ($G\!=\!8$) \\
Mish \\
FiLM: $\gamma \!\odot\! \mathbf{h} \!+\! \beta$ \\[1pt]
Conv1D ($k\!=\!5$) \\
GroupNorm ($G\!=\!8$) \\
Mish \\
\end{tabular}
};
\draw[thick, rounded corners=3pt, ->] ([xshift=-1.2cm]detail.north) -- ++(-0.3,0) |- ([xshift=-1.2cm]detail.south) node[midway, left, font=\scriptsize] {$+$};

\end{tikzpicture}%
}
\caption{\textbf{Overview of Ada3Drift.} \textit{Left}: the timestep-free 1D U-Net maps Gaussian noise $\mathbf{z}$ to action trajectories $\hat{\mathbf{x}}$. The encoder (blue) downsamples the temporal dimension, while the decoder (teal) upsamples with skip connections. The bottom row (training only) shows the drifting field loss on distance-rescaled actions $\tilde{\mathbf{x}}$: vectors aggregated across temperatures and combined with MSE loss via sigmoid scheduling. \textit{Right}: the 3D observation encoder extracts a global conditioning vector $\mathbf{g}$, which modulates every residual block via FiLM. Inference uses an observation encoder and a 1-NFE generator.}
\label{fig:architecture}
\end{figure*}

\section{Related Work}

\subsection{Visuomotor Policy Learning}
Visuomotor policy learning has evolved from early behavior cloning of image-to-action mappings~\cite{jang2022bcz, florence2022implicit} to transformer-based architectures with language conditioning~\cite{brohan2023rt1roboticstransformerrealworld, zitkovich2023rt2} and multi-task pretraining~\cite{octo2024, black2024pi0, Wang2024ScalingPL}. Action distributions can be modeled through latent-variable policies~\cite{zhao2023learning}, energy-based models~\cite{florence2022implicit}, or diffusion~\cite{chi2023diffusion}. Behavior Transformers~\cite{shafiullah2022bet} combine discrete action modes with continuous offsets. RoboMimic~\cite{mandlekar2022whatmatters} highlights sensitivity to demonstration quality and checkpoint selection, motivating matched data and evaluation protocols.
To address depth ambiguity in 2D policies, explicit 3D representations include voxel grids~\cite{shridhar2022peract}, neural feature fields~\cite{ze2023gnfactor, Gervet2023Act3D3F}, and point clouds~\cite{goyal2024rvt, ze20243d}. Recent work further explores hierarchical semantic fields for pose-aware manipulation~\cite{xu2026hero} and pretrained 3D geometric priors for joint action and future-geometry prediction in RGB-based bimanual manipulation~\cite{xu2026gap}. These developments make observation representation and action modeling complementary design choices. We focus on training objectives for one-step action generation from point clouds, keeping the observation encoder fixed in our controlled comparisons.

\subsection{Diffusion Models in Robotics}
Diffusion models~\cite{ho2020denoising} are now a dominant paradigm for multimodal action modeling. Diffusion Policy~\cite{chi2023diffusion} casts policy learning as conditional denoising, and follow-ups extend this with 3D point cloud conditioning (DP3~\cite{ze20243d}, evaluated on Meta-World~\cite{mclean2025metaworld} and Adroit~\cite{Rajeswaran-RSS-18}), hierarchical sub-goal diffusion~\cite{ma2024hierarchical}, large-scale foundation policies (Octo~\cite{octo2024}, RDT-1B~\cite{rdt2024}), and trajectory-level motion planning~\cite{carvalho2023motion}. All of them rely on iterative denoising (typically 10--100 NFE). Accelerated samplers like DDIM~\cite{song2021denoising} reduce but do not eliminate this cost, leaving a bottleneck for \mbox{real-time, high-frequency control.} This motivates treating sampling cost as a policy design constraint, alongside the quality of the generated action chunks.

\subsection{Efficient One-Step Policy Generation}
One-step policies replace iterative sampling through teacher distillation~\cite{prasad2024consistency, wang2024onestep} or direct training with specialized objectives~\cite{geng2025meanflowsonestepgenerative, zhang2025flowpolicy, mp1}. To preserve action structure, subsequent work improves visual representations~\cite{Zou2025DM1MW}, temporal consistency~\cite{Su2025FreqPolicyEF}, or latent mode modeling~\cite{Zhai2025VFPVF}. Reinforcement learning provides additional reward supervision~\cite{Wang2025OneStepGP, Zou2026OneSI}. These approaches improve what the generator learns while retaining efficient inference. A complementary question is how different supervision signals should be combined over the course of training.

Drifting Models~\cite{deng2026generativemodelingdrifting} instead refine the generated distribution during training through attraction and repulsion, and already demonstrate one-step robotic control. We build on this mechanism and study when to combine it with paired regression. Regression preserves observation--action correspondence, while our drifting field regularizes actions across a minibatch. Following curriculum learning~\cite{bengio2009curriculum}, we progressively change their weights, rather than replacing inputs as in scheduled sampling~\cite{bengio2015scheduled}. Our focus is this training schedule under limited demonstrations, with the original drifting field and multi-temperature aggregation retained.

\section{Method}
\label{sec:method}

\noindent We present \textbf{Ada3Drift}, a single-step 3D visuomotor policy that combines paired regression and training-time distribution refinement under limited demonstrations. We first discuss the speed--fidelity trade-off in robotic action spaces (\cref{sec:prelim}), then present our training-time refinement framework (\cref{sec:drifting}), and detail the architecture (\cref{sec:architecture}).

\subsection{The Speed-Fidelity Trade-off in Policy Generation}
\label{sec:prelim}

\noindent Diffusion-based policies~\cite{chi2023diffusion, ze20243d} generate action chunks through iterative denoising. Repeated network evaluations increase inference latency, motivating one-step alternatives. This computational advantage alone does not determine a generator's ability to represent multiple modes.

\noindent Flow Matching (FM)~\cite{lipman2023flow} learns a time-dependent velocity field. Omitting observation conditioning for clarity, given noise $\mathbf{z} \sim p_0$ and target trajectory $\mathbf{a} \sim p_1$, the conditional flow is:
\begin{equation}
    \psi_t(\mathbf{z} | \mathbf{a}) = (1 - t)\mathbf{z} + t\mathbf{a}, \quad t \in [0, 1],
\end{equation}
with conditional vector field $u_t(\mathbf{x} | \mathbf{a}) = \mathbf{a} - \mathbf{z}$. A network $v_\theta$ regresses this field:
\begin{equation}
    \mathcal{L}_{\text{FM}} = \mathbb{E}_{t, \mathbf{z}, \mathbf{a}} \left[ \| v_\theta(\psi_t(\mathbf{z} | \mathbf{a}), t) - (\mathbf{a} - \mathbf{z}) \|^2 \right].
\end{equation}
\noindent The optimum is a conditional expectation of velocities, which defines the marginal flow field rather than an averaged final action. Its trajectories can be curved even when the conditional paths are straight. A single Euler step may therefore incur large integration error. MeanFlow~\cite{geng2025meanflowsonestepgenerative} instead learns average velocity over a time interval to support one-step generation. One-step objectives do not universally lose multimodality.

\noindent\textbf{Why action averaging can be problematic.}
For a given observation, demonstrations may contain distinct valid strategies, such as passing an obstacle on either side. Their arithmetic average can intersect the obstacle. Paired MSE with independently sampled noise can favor this conditional mean and ignore the noise input. This motivates combining paired supervision with distribution-level regularization~\cite{chi2023diffusion}, although neither action feasibility nor conditional mode coverage follows from the loss alone. Our experiments evaluate task success, not a guarantee of multimodal recovery.

\subsection{Training-Time Refinement via Drifting Fields}
\label{sec:drifting}

\noindent\textbf{From compute asymmetry to training-time drifting.}
Offline training permits repeated optimization without adding sampling steps at deployment. Following Drifting Models~\cite{deng2026generativemodelingdrifting}, we use a sample-dependent field to define detached regression targets during training, while the generator maps noise and observations directly to an action chunk at inference.

\noindent The field is computed from generated and demonstrated actions, rather than predicted by a separate learned network. Attraction and repulsion provide batch-level distribution information. They do not group samples by identical observations. We therefore combine this inherited mechanism with paired MSE and investigate when to introduce it through a curriculum.

\noindent\textbf{Drifting field computation.}
For one task, a minibatch contains paired observations and action chunks $(\mathbf{g}_i,\mathbf{y}_i^+)$. One independent noise sample per observation produces $\mathbf{x}_i=f_\theta(\mathbf{z}_i,\mathbf{g}_i)$. Actions are dataset-normalized and flattened to $D=D_aH$ dimensions. Let $S$ be the mean pairwise distance in $\{\mathbf{x}_i,\mathbf{y}_i^+\}$ divided by $\sqrt D$. We use $\tilde{\mathbf{x}}=\mathbf{x}/\mathrm{sg}(S)$ and similarly scaled targets, with small numerical stabilizers. The pooled support $\{\tilde{\mathbf{y}}_j\}$ includes expert positives and generated negatives. Affinities are:
\begin{equation}
    A_{ij} = \sqrt{A_{ij}^{\text{row}} \cdot A_{ij}^{\text{col}}},
\end{equation}
where $A^{\text{row}}$ and $A^{\text{col}}$ are row-wise and column-wise softmax normalizations of pairwise distance logits:
\begin{equation}
\begin{aligned}
    A_{ij}^{\text{row}} &= \frac{\exp(-\|\tilde{\mathbf{x}}_i - \tilde{\mathbf{y}}_j\| / T)}{\sum_k \exp(-\|\tilde{\mathbf{x}}_i - \tilde{\mathbf{y}}_k\| / T)}, \\
    A_{ij}^{\text{col}} &= \frac{\exp(-\|\tilde{\mathbf{x}}_i - \tilde{\mathbf{y}}_j\| / T)}{\sum_k \exp(-\|\tilde{\mathbf{x}}_k - \tilde{\mathbf{y}}_j\| / T)}.
\end{aligned}
\end{equation}
\noindent Here $T=\tau\sqrt D$ is the effective kernel temperature. Row normalization includes both support sets. The controlled ablations mask each prediction's self-negative before normalization. Splitting $A$ into $A^+$ and $A^-$ gives $W^+_{ij}=A^+_{ij}\sum_k A^-_{ik}$ and $W^-_{ik}=A^-_{ik}\sum_j A^+_{ij}$. The field is:
\begin{equation}
    V(\tilde{\mathbf{x}}_i) = \underbrace{\sum_j W_{ij}^+ \tilde{\mathbf{y}}_j^+}_{\text{expert attraction}} - \underbrace{\sum_k W_{ik}^- \tilde{\mathbf{x}}_k}_{\text{generated repulsion}},
    \label{eq:drift_field}
\end{equation}
\noindent These cross-set weights follow Drifting Models~\cite{deng2026generativemodelingdrifting}. They favor nearby expert actions and discourage generated-sample concentration, without guaranteeing coverage of every mode. Since the comparisons span different observations, the field regularizes the task's action population rather than explicitly matching each observation's conditional distribution.

\noindent\textbf{Multi-scale field aggregation.}
We retain the original temperatures $\{\tau_l\}=\{0.02,0.05,0.2\}$ to combine interactions at different distances in normalized action space~\cite{deng2026generativemodelingdrifting}:
\begin{equation}
    V_{\text{total}}(\tilde{\mathbf{x}}) = \sum_{l} \frac{V_{\tau_l}(\tilde{\mathbf{x}})}{\lambda_{\tau_l}},
    \label{eq:multi_temp}
\end{equation}
\noindent where $\lambda_{\tau_l}=\sqrt{\mathbb{E}[\|V_{\tau_l}\|^2/D]+\epsilon}$ scales each field by its root-mean-square coordinate magnitude. The expectation is a minibatch average, and the scale is detached. Multiple temperatures and RMS normalization are inherited components. Together, these operations define the distribution-level signal whose influence is controlled by the loss schedule below.

\noindent\textbf{Sigmoid-scheduled loss transition.}
Paired regression supplies a direct observation--action learning signal, whereas drifting adds interactions across the minibatch. We hypothesize that learning the predictive mapping first provides a useful initialization for this regularizer. The combined objective is:
\begin{equation}
    \mathcal{L} = w_{\text{drift}}(e) \cdot \mathcal{L}_{\text{drift}} + w_{\text{mse}}(e) \cdot \operatorname{MSE}(\mathbf{x},\mathbf{y}^+),
    \label{eq:total_loss}
\end{equation}
\noindent where $\mathcal{L}_{\text{drift}}=\operatorname{MSE}(\tilde{\mathbf{x}},\mathrm{sg}(\tilde{\mathbf{x}}+V_{\text{total}}))$. MSE averages over samples and action coordinates. The paired term uses dataset-normalized actions before distance rescaling, while the drift term uses $\tilde{\mathbf{x}}$. With $\sigma$ denoting the logistic sigmoid, weights follow:
\begin{equation}
    w_{\text{drift}}(e) = \sigma\!\left(\frac{e - e_{\text{mid}}}{k \cdot E}\right), \quad
    w_{\text{mse}}(e) = 1 - w_{\text{drift}}(e),
\end{equation}
\noindent with $e_{\text{mid}}=0.7E$ and $k=0.05$. Early epochs emphasize paired regression, and later epochs emphasize drifting (\cref{fig:drifting}~(b)). This preset curriculum controls the transition by epoch rather than observed convergence. We compare it with other objectives and schedules experimentally. Because early MSE can reduce sensitivity to noise, conditional diversity after the transition remains a separate property to evaluate. The schedule changes only the training objective. Both terms update the same observation-conditioned generator, whose architecture remains fixed throughout training.

\subsection{Architecture}

\newcommand{\std}[1]{\text{\scriptsize\textcolor{gray}{\ensuremath{\pm#1}}}}

\begin{table*}[t]
    \centering
    \caption{\textbf{Comparison on Adroit~\cite{Rajeswaran-RSS-18} and Meta-World~\cite{mclean2025metaworld} with 10 expert demonstrations.} We evaluate against \textbf{image-based} ($^*$, cited from~\cite{zhang2025flowpolicy}) and \textbf{3D point cloud-based} baselines (reproduced). Best 3D means are \textbf{bold}, and second-best means are \textnormal{\textit{italicized}}.}
    \label{tab:success}
    
    \resizebox{\textwidth}{!}{%
    \begin{tabular}{l c ccc cccc c}
        \toprule
        \multirow{2}{*}{\textbf{Methods}} & \multirow{2}{*}{\textbf{NFE}} & \multicolumn{3}{c}{\textbf{Adroit}} & \multicolumn{4}{c}{\textbf{Meta-World}} & \multirow{2}{*}{\textbf{Avg.}} \\
        
        \cmidrule(lr){3-5} \cmidrule(lr){6-9}
         & & \textit{Hammer} & \textit{Door} & \textit{Pen} & Easy & Medium & Hard & V. Hard & \\
        \midrule
        
        DP$^*$~\cite{chi2023diffusion} & 10 & 16\std{10} & 34\std{11} & 13\std{2} & 50.7\std{6.1} & 11.0\std{2.5} & 5.3\std{2.5} & 22.0\std{5.0} & 35.2\std{5.3} \\
        AdaFlow$^*$~\cite{hu2024adaflow} & -- & 45\std{11} & 27\std{6} & 18\std{6} & 49.4\std{6.8} & 12.0\std{5.0} & 5.8\std{4.0} & 24.0\std{4.8} & 35.6\std{6.1} \\
        CP$^*$~\cite{prasad2024consistency} & 1 & 45\std{4} & 31\std{10} & 13\std{6} & 69.3\std{4.2} & 21.2\std{6.0} & 17.5\std{3.9} & 30.0\std{4.9} & 50.0\std{4.7} \\
        
        \midrule
        DP3~\cite{ze20243d} & 10 & \textit{88.7}\std{1.2} & \textit{64.2}\std{6.5} & 59.7\std{3.1} & 85.5\std{1.9} & \textit{64.0}\std{3.5} & 58.6\std{1.4} & 71.5\std{2.5} & 77.2\std{2.4} \\
        Simple DP3~\cite{ze20243d} & 10 & 86.0\std{2.4} & 61.0\std{4.5} & \textit{61.0}\std{2.8} & \textit{85.9}\std{1.9} & \textbf{64.3}\std{3.7} & \textit{61.1}\std{3.4} & 70.7\std{2.5} & 77.5\std{2.7} \\
        FlowPolicy~\cite{zhang2025flowpolicy} & 1 & 77.0\std{2.2} & 61.2\std{4.7} & 58.0\std{4.1} & 84.3\std{2.6} & 62.7\std{3.2} & \textit{61.1}\std{3.5} & 71.2\std{2.7} & 76.2\std{3.0} \\
        MP1~\cite{mp1} & 1 & 84.3\std{3.1} & \textit{64.2}\std{3.5} & 57.7\std{4.5} & 85.8\std{1.5} & 62.2\std{7.5} & \textbf{62.3}\std{2.0} & \textbf{74.4}\std{4.3} & \textit{77.8}\std{3.2} \\
        \rowcolor{gray!10}
        \textbf{Ada3Drift (Ours)} & \textbf{1} & \textbf{90.3}\std{0.9} & \textbf{65.0}\std{3.6} & \textbf{63.3}\std{3.1} & \textbf{86.7}\std{1.7} & 62.3\std{2.7} & 60.7\std{6.2} & \textit{72.7}\std{4.5} & \textbf{78.2}\std{2.8} \\
        \bottomrule
    \end{tabular}%
    }
    \begin{flushleft}
        \scriptsize $^*$ Results are cited from FlowPolicy~\cite{zhang2025flowpolicy}. All other baselines are reproduced on a single RTX 4090D.
    \end{flushleft}
\end{table*}

\label{sec:architecture}

\noindent Ada3Drift takes the robot's proprioceptive state $\mathbf{s}_t$ and a 3D point cloud $\mathbf{P}_t$ as input, and outputs a horizon of $H$ future actions $\hat{\mathbf{a}}_{t:t+H}$ in a forward pass. The architecture comprises a 3D observation encoder, a timestep-free action generator, and a receding-horizon execution scheme, as in \cref{fig:architecture}.

\noindent\textbf{3D Observation Encoder.}
We encode point cloud observations following the DP3 pipeline~\cite{ze20243d}. Training observations contain 512 points for Adroit and 1024 for Meta-World and RoboTwin. The Meta-World ablation rollout runner samples 512 points by Farthest Point Sampling (FPS). A PointNet backbone~\cite{qi2017pointnet} processes XYZ coordinates through shared MLPs with channels $[64,128,256]$, max-pooling, and a linear projection with LayerNorm to a 64-dimensional point-cloud feature. A separate MLP maps proprioception $\mathbf{s}_t$ to 64 dimensions. Concatenation yields $\mathbf{c}_t\in\mathbb{R}^{128}$, without a further projection. For $T_o=2$ observation steps, $\mathbf{g}=[\mathbf{c}_1;\mathbf{c}_2]\in\mathbb{R}^{256}$.

\noindent\textbf{Timestep-Free Action Generator.}
In diffusion and flow-based policies, the network is conditioned on a noise level or timestep $t$ via a sinusoidal embedding, because the same architecture must operate across all noise scales during denoising. Since Ada3Drift relocates all refinement to training and generates actions in a single pass, this conditioning is unnecessary. We remove the timestep embedding and its projection layers entirely, yielding a simpler architecture.

\noindent The generator is a 1D convolutional U-Net that maps Gaussian noise $\mathbf{z}\in\mathbb{R}^{D_a\times H}$ to an action trajectory, with $H=16$. An input projection produces 512 channels. Two encoder stages, each containing two residual blocks, increase channels to 1024 and 2048. A single strided convolution between these stages reduces the temporal length from 16 to 8. Two 2048-channel residual blocks form the bottleneck. Two decoder stages use concatenated encoder skips and reduce channels to 1024 and 512. A transposed convolution between them restores temporal length 16.

\noindent Each residual block contains two Conv1D layers (kernel size 5, padding 2) with GroupNorm ($G=8$ groups) and Mish activations, plus a residual shortcut (1$\times$1 convolution when channel dimensions differ, identity otherwise). The observation features modulate intermediate representations via FiLM conditioning~\cite{perez2018film} applied at \emph{every} residual block:
\begin{equation}
    \mathbf{h}' = \gamma(\mathbf{g}) \odot \text{Mish}(\text{GroupNorm}(\text{Conv}(\mathbf{h}))) + \beta(\mathbf{g}),
\end{equation}
\noindent where $\gamma, \beta \in \mathbb{R}^{C}$ are per-channel scale and bias vectors predicted from $\mathbf{g}$ through a Mish-activated linear layer. This conditioning injects observation features at each temporal resolution. A final Conv1D block followed by a $1 \times 1$ convolution projects the decoder output back to the action dimension $D_a$.

\noindent The resulting network is shared between training and deployment. Drifting targets are needed only to compute the training loss, not to generate actions at test time.

\noindent\textbf{Receding-Horizon Execution.}
At test time, action generation requires exactly one forward pass: sample $\mathbf{z} \sim \mathcal{N}(\mathbf{0}, \mathbf{I})$, compute $\hat{\mathbf{a}} = f_\theta(\mathbf{z}, \mathbf{g})$, and extract the executable chunk $\hat{\mathbf{a}}_{T_o-1 : T_o-1+N_a}$ with execution length $N_a = 8$ from the full $H = 16$ horizon. This receding-horizon scheme allows the policy to correct for prediction drift by replanning every $N_a$ steps. The entire pipeline yields \textbf{1 NFE} per action chunk, compared to 10--100 NFE for diffusion-based policies, reducing the number of sequential generator evaluations. End-to-end control latency also includes observation processing and action execution.

\begin{table*}[t]
    \centering
    \caption{\textbf{RoboTwin~\cite{mu2025robotwin} single-arm results.} All methods use 3D point clouds. Best means in \textbf{bold}, second-best in \textnormal{\textit{italics}}.}
    \label{tab:robotwin}
    \setlength{\tabcolsep}{3.5pt}
    \resizebox{\textwidth}{!}{%
    \begin{tabular}{l c cccccccc c}
        \toprule
        \multirow{2}{*}{\textbf{Methods}} & \multirow{2}{*}{\textbf{NFE}} & \textit{Beat} & \textit{Click} & \textit{Open} & \textit{Place} & \textit{Place} & \textit{Stack} & \textit{Stack} & \textit{Turn} & \multirow{2}{*}{\textbf{Avg.}} \\
         & & \textit{Block} & \textit{Alarm} & \textit{Microwave} & \textit{Cup} & \textit{Phone} & \textit{Blocks} & \textit{Bowls} & \textit{Switch} & \\
        \midrule
        DP3~\cite{ze20243d} & 10 & 63.7 \std{2.6} & 83.0 \std{2.2} & \textit{95.0} \std{2.2} & 58.7 \std{5.4} & 43.7 \std{8.2} & \textbf{30.0} \std{1.6} & 82.0 \std{6.2} & 44.3 \std{4.5} & 62.5 \\
        FlowPolicy~\cite{zhang2025flowpolicy} & 1 & 56.3 \std{1.2} & \textit{97.0} \std{0.0} & 66.3 \std{3.7} & 76.7 \std{1.2} & \textbf{55.5} \std{5.5} & 10.7 \std{2.4} & 61.7 \std{1.7} & 42.7 \std{2.1} & 58.4 \\
        MP1~\cite{mp1} & 1 & \textit{79.0} \std{1.6} & 95.0 \std{1.4} & 94.7 \std{1.2} & \textit{81.0} \std{3.7} & 41.0 \std{2.2} & \textit{23.7} \std{1.2} & \textit{86.3} \std{2.1} & \textit{46.0} \std{1.4} & \textit{68.3} \\
        \rowcolor{gray!10}
        \textbf{Ada3Drift (Ours)} & \textbf{1} & \textbf{81.3} \std{3.4} & \textbf{97.7} \std{1.2} & \textbf{97.7} \std{1.2} & \textbf{86.7} \std{0.5} & \textit{46.7} \std{5.4} & 21.0 \std{2.8} & \textbf{88.3} \std{1.7} & \textbf{50.0} \std{4.3} & \textbf{71.2} \\
        \bottomrule
    \end{tabular}%
    }
\end{table*}

\begin{table*}[t]
\centering
\caption{\textbf{Training objective and curriculum ablation.} Ten demonstrations, 12 tasks, seeds 0--2. Success (\%): mean $\pm$ SD over seeds, with equal task weights per cell and top-5 checkpoint scores. Best/second means: \textbf{bold}/\textnormal{\textit{italics}}.}
\label{tab:ablation}
\setlength{\tabcolsep}{2pt}
\small
\begin{tabular*}{\linewidth}{@{\extracolsep{\fill}}lcccccccc@{}}
\toprule
\multirow{2}{*}{Training variant} & \multicolumn{3}{c}{Adroit} & \multicolumn{4}{c}{Meta-World} & \multirow{2}{*}{Avg.} \\
\cmidrule(lr){2-4}\cmidrule(lr){5-8}
 & \textit{Hammer} & \textit{Door} & \textit{Pen} & Easy & Medium & Hard & V. Hard & \\
\midrule
Direct Drifting & 2.3$\std{4.0}$ & 3.7$\std{3.1}$ & 13.2$\std{6.8}$ & 50.7$\std{18.2}$ & 4.7$\std{3.0}$ & 11.2$\std{5.1}$ & 0.3$\std{0.6}$ & 17.0$\std{5.4}$ \\
Regression only & 82.7$\std{6.7}$ & \textit{64.2}$\std{9.9}$ & \textit{59.8}$\std{12.4}$ & 74.2$\std{1.3}$ & 35.2$\std{3.4}$ & \textbf{69.0}$\std{1.0}$ & 44.3$\std{6.4}$ & 60.5$\std{2.9}$ \\
Fixed mixture & 68.3$\std{6.8}$ & 13.0$\std{2.6}$ & 48.8$\std{7.0}$ & 66.6$\std{1.4}$ & 16.8$\std{2.5}$ & 32.0$\std{7.7}$ & 14.0$\std{2.3}$ & 38.0$\std{2.0}$ \\
Hard switch & 85.7$\std{9.5}$ & 60.2$\std{4.8}$ & 55.8$\std{9.8}$ & 74.3$\std{1.2}$ & \textit{38.7}$\std{1.5}$ & 67.3$\std{1.0}$ & \textbf{47.3}$\std{3.2}$ & 60.9$\std{1.7}$ \\
Linear curriculum &  \textit{87.3}$\std{11.2}$ & 61.2$\std{7.0}$ & 57.8$\std{8.5}$ & \textit{74.8}$\std{1.1}$ & 37.3$\std{3.3}$ & 68.7$\std{1.3}$ & 45.3$\std{0.8}$ & \textit{61.1}$\std{1.6}$ \\
\oursrowbackground
\textbf{Ada3Drift (Ours)} & \textbf{90.3}\std{0.9} & \textbf{65.0}\std{3.6} & \textbf{63.3}\std{3.1} & \textbf{75.3}$\std{1.2}$ & \textbf{39.5}$\std{6.4}$ & \textit{68.8}$\std{1.0}$ & \textit{45.7}$\std{5.2}$ & \textbf{62.7}$\std{2.0}$ \\
\bottomrule
\end{tabular*}
\end{table*}


\section{Experiments}


\subsection{Experimental Setup}

\noindent\textbf{Benchmarks.}
Following FlowPolicy~\cite{zhang2025flowpolicy}, we first evaluate on \textbf{Adroit}~\cite{Rajeswaran-RSS-18} (3 dexterous manipulation tasks with a 24-DoF Shadow Hand) and \textbf{Meta-World}~\cite{mclean2025metaworld} (34 tasks on a Sawyer arm, grouped into Easy (21), Medium (4), Hard (4), and Very Hard (5)) with 10 expert demonstrations per task.
To further assess 3D perception capability, we select 8 representative tasks from the \textbf{RoboTwin}~\cite{mu2025robotwin} simulation platform on the Agilx Cobot Magic robot, each requiring precise spatial reasoning from point cloud observations.
Finally, following the real-world evaluation protocol of MP1~\cite{mp1}, we design 5 manipulation tasks on a physical Agilx Cobot Magic robot inspired by RoboTwin task designs, each trained from 20 expert demonstrations and evaluated over 20 trials.

\noindent\textbf{Baselines.}
We compare against two categories of methods:
(i) \textit{Image-based}: Diffusion Policy (DP)~\cite{chi2023diffusion}, AdaFlow~\cite{hu2024adaflow}, and Consistency Policy (CP)~\cite{prasad2024consistency}, with results cited from~\cite{zhang2025flowpolicy}, and
(ii) \textit{3D point cloud-based}: DP3~\cite{ze20243d}, Simple DP3~\cite{ze20243d}, FlowPolicy~\cite{zhang2025flowpolicy}, and MP1~\cite{mp1}, all reproduced under identical settings.
Among these, DP, DP3, and Simple DP3 use iterative generation (10 NFE), while CP, FlowPolicy, MP1, and Ada3Drift achieve single-step generation (1 NFE).

\noindent\textbf{Implementation Details.}
We use a compact observation consisting of the robot's proprioceptive state and a 3D point cloud, downsampled via FPS to 512 points (Adroit) or 1024 points (Meta-World, RoboTwin). The observation horizon is 2 steps. We use 10 expert demonstrations per task for Adroit and Meta-World. All models are trained on a single NVIDIA RTX 4090D with AdamW ($\text{lr}=10^{-4}$, batch size 128). We report success rates averaged over 3 seeds, evaluating checkpoints every 200 epochs and reporting the mean of the top-5 checkpoints per seed.

\subsection{Main Results}

\noindent\textbf{Adroit and Meta-World.}
\cref{tab:success} presents the comparison on Adroit and Meta-World. Among one-step methods, Ada3Drift achieves the highest average success rate (78.2\%), outperforming both FlowPolicy (76.2\%) and MP1 (77.8\%). Notably, Ada3Drift surpasses MP1 on 5 out of 7 task categories, with particularly strong gains on Adroit \textit{Hammer} (+6.0 percentage points), \textit{Pen} (+5.6 points), and Meta-World Easy (+0.9 points). Ada3Drift also matches or exceeds the multi-step DP3 baseline (77.2\%, 10 NFE) using one-tenth as many function evaluations, showing competitive average task success with fewer sequential generator evaluations.
Ada3Drift underperforms MP1 on Meta-World Hard (60.7\% vs.\ 62.3\%) and Very Hard (72.7\% vs.\ 74.4\%). The aggregate scores do not identify whether these differences arise from action ambiguity, optimization, or other task-specific factors.

\noindent\textbf{RoboTwin.}
\cref{tab:robotwin} evaluates all methods on 8 manipulation tasks from the RoboTwin benchmark, which requires precise spatial reasoning from point cloud observations. Ada3Drift achieves the highest average success rate of 71.2\%, exceeding the reported means of DP3, FlowPolicy, and MP1. Ada3Drift ranks first on 6 out of 8 tasks, with particularly large margins on \textit{Place Cup} (86.7\% vs.\ 81.0\% for MP1, +5.7 points), \textit{Turn Switch} (50.0\% vs.\ 46.0\%, +4.0 points), and \textit{Open Microwave} (97.7\% vs.\ 94.7\%, +3.0 points). Notably, Ada3Drift surpasses the multi-step DP3 by 8.7 percentage points on average despite using only a single forward pass, showing that iterative inference is not required to attain this average on the evaluated tasks. FlowPolicy has lower scores on \textit{Open Microwave} (66.3\%) and \textit{Stack Bowls} (61.7\%). These differences do not establish mode averaging as their cause.

\subsection{Analysis and Ablation}

\noindent\textbf{Training objective and curriculum.}
\cref{tab:ablation} uses three Adroit and nine Meta-World tasks: \textit{reach}, \textit{door-open}, and \textit{plate-slide} (Easy), \textit{peg-insert-side} and \textit{soccer} (Medium), \textit{push} and \textit{pick-place} (Hard), and \textit{stick-pull} and \textit{shelf-place} (Very Hard). All variants share architecture, data, 3000 epochs, and evaluation. Drift weights are 0.3 (fixed), switched at $0.7E$ (hard), or ramped over $[0.6E,0.8E]$ (linear), approximately matching sigmoid's cumulative weight. All drifting variants retain multi-temperature normalization. Meta-World ablations use 1024/512 points for training/rollouts. Scores average the top five of 15 evaluations of 20 episodes, without independent final testing. Sigmoid has the highest displayed average (62.7\%). Average SDs await matching per-seed verification.

\noindent\textbf{Demonstration budget.}
\cref{tab:ablation_demos} uses the same 12 tasks, training seed 0, and nested subsets (subset seed 42), with normalization fitted only on training episodes. Sigmoid has the highest displayed mean at each budget: 46.8/62.2/73.4\%, exceeding the strongest alternatives by 0.6/0.4/0.4 points. All methods improve with more demonstrations, but equal epochs imply unequal optimizer steps across budgets. One seed cannot establish robustness.

\begin{table}[t]
\centering
\caption{\textbf{Demonstration-count ablation.}\\Mean success (\%), 12 tasks, seed 0.}
\label{tab:ablation_demos}
\small
\begin{tabular*}{\linewidth}{@{\extracolsep{\fill}}lccc@{}}
\toprule
\multirow{2}{*}{Method} & \multicolumn{3}{c}{Demonstrations per task} \\
\cmidrule(lr){2-4}
 & 5 & 10 & 20 \\
\midrule
Direct Drifting & 9.2 & 10.8 & 21.6 \\
Regression only & 43.5 & \textit{61.8} & 72.6 \\
Fixed mixture & 21.4 & 36.2 & 47.0 \\
Hard switch & \textit{46.2} & 61.5 & \textit{73.0} \\
\oursrowbackground
\textbf{Ada3Drift (Ours)} & \textbf{46.8} & \textbf{62.2} & \textbf{73.4} \\
\bottomrule
\end{tabular*}
\end{table}

\begin{figure}[t]
\centering
\includegraphics[width=\linewidth]{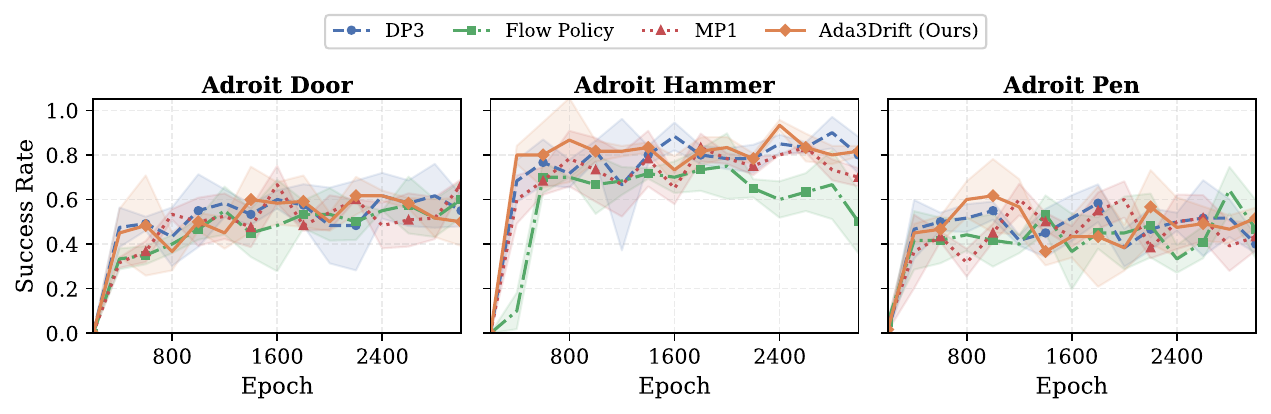}
\par\nointerlineskip
\caption{\textbf{Training curves on Adroit.} Success rate (mean $\pm$ std over 3 seeds) versus training epoch.}
\label{fig:training_curves}
\end{figure}

\noindent\textbf{Training dynamics.}
\cref{fig:training_curves} plots success rates over training epochs on three Adroit tasks. On \textit{Hammer}, Ada3Drift leads from epoch 600 and finishes above 90\%. On \textit{Door}, the single-step methods converge to similar performance, with less variation across seeds for Ada3Drift. On \textit{Pen}, Ada3Drift separates from FlowPolicy and MP1 later in training. These curves compare complete policies. The matched comparisons in \cref{tab:ablation} vary the training objective under the same architecture and data.

\Needspace{10\baselineskip}
\noindent\textbf{Inference speed.}
We evaluate action-chunk inference on Adroit \textit{Door} using a single RTX 4090D (batch size 1).

\begingroup
\setlength{\columnsep}{10pt}
\begin{wraptable}{r}{0.54\columnwidth}
    \vspace{-\intextsep}
    \centering
    \setlength{\tabcolsep}{1.5pt}
    \small
    \begin{tabular*}{\linewidth}{@{\extracolsep{\fill}}lccc@{}}
        \toprule
        Method & NFE & Hz $\uparrow$ & ms $\downarrow$ \\
        \midrule
        DP3 & 10 & 18.7 & 53.4 \\
        FlowPolicy & 1 & \textbf{274.4} & \textbf{3.6} \\
        MP1 & 1 & \textit{249.5} & \textit{4.0} \\
        \oursrowbackground
        Ada3Drift (Ours) & 1 & 233.9 & 4.3 \\
        \bottomrule
    \end{tabular*}
\end{wraptable}

\noindent At 233.9\,Hz, Ada3Drift takes 4.3\,ms per chunk, a $12.5\times$ speedup over DP3 with 10 DDIM steps. FlowPolicy and MP1 reach 274.4\,Hz and 249.5\,Hz, respectively. All three one-step methods use a single generator evaluation per chunk. These are action-chunk inference rates. End-to-end control also includes perception, communication, and execution of eight actions per chunk.\par
\endgroup



\begin{figure*}[t]
\centering
\includegraphics[width=\linewidth]{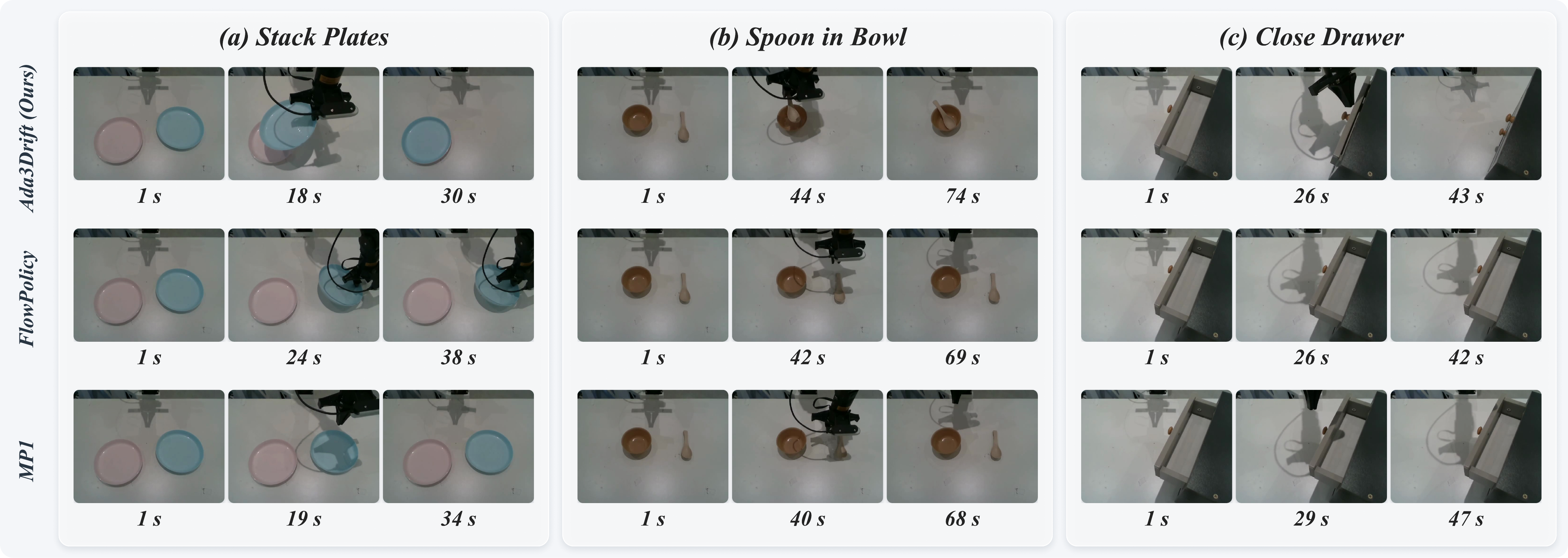}
\par\nointerlineskip
\caption{\textbf{Selected real-world rollouts on three tasks.} Rows compare Ada3Drift, FlowPolicy, and MP1. Each sequence shows early, intermediate, and late frames from one trial. Timestamps identify frames within each source video and do not measure synchronized execution times.}
\label{fig:realworld-rollouts}
\end{figure*}

\Needspace{4\baselineskip}
\noindent\textbf{Interpreting the objective comparisons.}
In \cref{tab:ablation}, regression alone reaches 60.5\%, compared with 17.0\% for direct drifting. This supports paired supervision under the tested protocol, without implying that Drifting Models generally fail at robotic control. The fixed mixture reaches 38.0\%, versus 60.9\%, 61.1\%, and 62.7\% for hard, linear, and sigmoid schedules. Approximately matched cumulative drift weights help compare when the regularizer is applied, but do not equate gradient magnitudes or optimization trajectories.

\noindent Sigmoid exceeds regression by 2.2 points and the strongest alternative schedule by 1.6 points on average. However, regression leads on Hard and hard switching on Very Hard, so the advantage is not uniform. The smaller 0.4--0.6-point margins in \cref{tab:ablation_demos} motivate repeated trials. Neither table measures conditional action diversity or directly establishes mode recovery. The tables probe sensitivity to the objective and demonstration count separately. Different seed protocols mean that the budget study does not replicate the three-seed objective comparison across independent training runs. Next, we assess the complete policy on hardware across five tasks.

\Needspace*{6\baselineskip}
\subsection{Real-World Experiments}

\begingroup
\setlength{\columnsep}{10pt}
\begin{wrapfigure}{r}{0.42\columnwidth}
\vspace{-\intextsep}
\centering
\includegraphics[width=\linewidth]{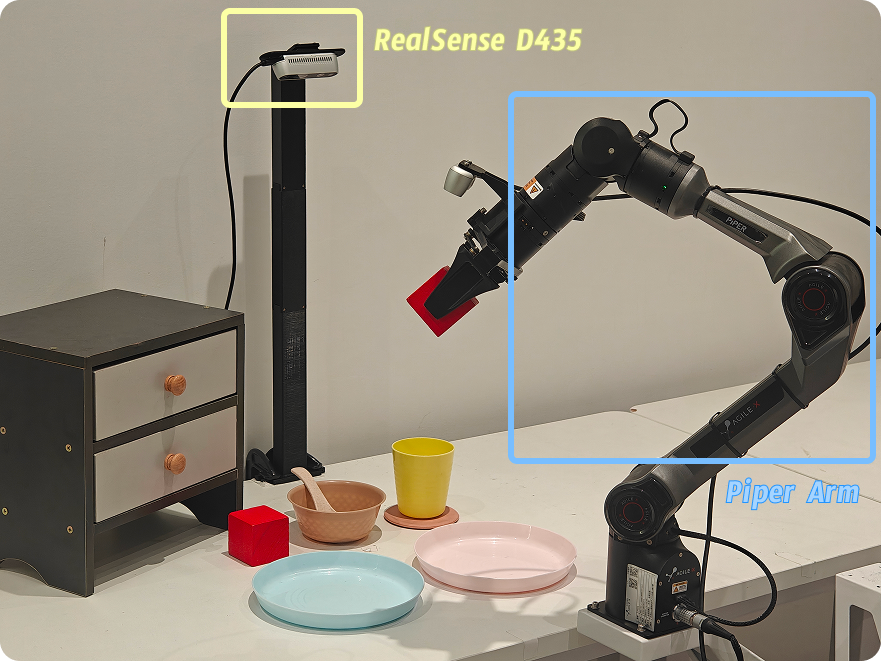}
\par\nointerlineskip
\caption{\textbf{Real-world setup.} Camera (yellow) and arm (blue).}
\label{fig:realworld}
\end{wrapfigure}

\noindent We deploy~Ada3Drift on one arm of the Agilx Cobot Magic robot, using 20 teleoperated expert demonstrations and 20 evaluation trials per task. We evaluate five manipulation tasks spanning placement, stacking, and pushing.
In \textit{Cup on Coaster}, the robot places a cup precisely onto a coaster.
\textit{Stack Blocks} requires stacking two same-color blocks, while \textit{Stack Plates} requires stacking two plates.
In \textit{Close~Drawer}, the robot pushes a drawer closed.
\textit{Spoon in Bowl} requires placing a spoon inside a bowl.
These tasks require the policy to use point cloud observations to guide object manipulation toward the target configuration.\par
\endgroup

\noindent\textbf{Overall performance.}
\cref{tab:realworld} shows that Ada3Drift achieves the highest mean success rate, 79\%, exceeding MP1, DP3, and FlowPolicy by 10, 11, and 22 percentage points, respectively. It ranks first on four tasks and ties DP3 on \textit{Close Drawer}. Against the two single-step baselines, its gains are positive on every task, showing that the average improvement is distributed across the evaluated manipulation primitives.

\noindent\textbf{Task-level gains.}
Compared with MP1, Ada3Drift gains 15 percentage points on \textit{Stack Blocks} and 10 on \textit{Cup on Coaster}, \textit{Stack Plates}, and \textit{Spoon in Bowl}. Against the strongest baseline per task, the margins are 10 points on \textit{Cup on Coaster} and \textit{Stack Plates}, and 5 on \textit{Stack Blocks} and \textit{Spoon in Bowl}. On \textit{Close Drawer}, its 90\% success rate matches DP3 and exceeds MP1 (85\%) and FlowPolicy (80\%). Relative to DP3, gains are 30 points on \textit{Stack Plates}, 15 on \textit{Cup on Coaster}, and 5 on both \textit{Stack Blocks} and \textit{Spoon in Bowl}. This uneven profile favors placement and stacking while retaining competitive pushing performance, and motivates reading the average together with individual task results.

\begin{table}[t]
    \centering
    \caption{\textbf{Real-world success (\%).} 20 demos and 20 trials/task. Best/second-best: \textbf{bold}/\textnormal{\textit{italics}}.}
    \label{tab:realworld}
    \setlength{\tabcolsep}{4pt}
    \resizebox{\linewidth}{!}{%
    \begin{tabular}{l ccccc c}
        \toprule
        \multirow{2}{*}{\textbf{Methods}} & \textit{Cup on} & \textit{Stack} & \textit{Stack} & \textit{Close} & \textit{Spoon in} & \multirow{2}{*}{\textbf{Avg.}} \\
         & \textit{Coaster} & \textit{Blocks} & \textit{Plates} & \textit{Drawer} & \textit{Bowl} & \\
        \midrule
        DP3~\cite{ze20243d} & 70 & \textit{55} & 60 & \textbf{90} & \textit{65} & 68 \\
        FlowPolicy~\cite{zhang2025flowpolicy} & \textit{75} & 25 & 55 & 80 & 50 & 57 \\
        MP1~\cite{mp1} & \textit{75} & 45 & \textit{80} & \textit{85} & 60 & \textit{69} \\
        \rowcolor{gray!10}
        \textbf{Ada3Drift (Ours)} & \textbf{85} & \textbf{60} & \textbf{90} & \textbf{90} & \textbf{70} & \textbf{79} \\
        \bottomrule
    \end{tabular}%
    }
\end{table}

\noindent\textbf{Remaining challenges.}
\textit{Stack Blocks} is still the least successful Ada3Drift task at 60\%, and \textit{Spoon in Bowl} reaches 70\%, compared with 90\% on \textit{Stack Plates} and \textit{Close Drawer}. These results show that success varies with object geometry and the type of manipulation. With 20 trials per task, a 5-point gap represents one successful trial. The individual margins should therefore be interpreted alongside the consistent ranking across tasks and the limited evaluation sample size per task.

\noindent\textbf{Grasping and placement behavior.}
To connect task-level results with visible behavior, \cref{fig:realworld-rollouts} shows selected executions from initial contact through object placement. On \textit{Stack Plates}, Ada3Drift transfers the blue plate onto the pink plate and withdraws the gripper. FlowPolicy leaves the plates separate, whereas MP1 moves the blue plate but releases it beside the target. The latter example distinguishes successful object movement from successful placement. On \textit{Spoon in Bowl}, Ada3Drift places the spoon inside the bowl, while both baseline clips end with the spoon outside despite the gripper approaching it. These final object configurations help interpret the placement-related differences in the table beyond the binary success outcome for each selected robot execution.

\medskip
\noindent\textbf{Pushing behavior.}
On \textit{Close Drawer}, Ada3Drift closes the drawer and withdraws, while it remains open in the selected FlowPolicy and MP1 clips. Timestamps locate frames within separate recordings and do not support comparisons of task-completion time or inference speed. The clips localize visible failures without quantifying their frequency. A systematic failure taxonomy requires labeling all evaluation trials. Moreover, five tasks on one platform do not establish robustness to new objects, camera viewpoints, or hardware. These factors require separate controlled tests under explicitly specified evaluation conditions and a fixed demonstration budget across methods. Changing one factor at a time would clarify its effect.

%

\Needspace{15\baselineskip}
\section{Conclusion}
\label{sec:conclusion}

\textbf{Ada3Drift} combines paired action regression with training-time drifting for one-step point-cloud-conditioned control. Building on Drifting Models, its sigmoid curriculum shifts emphasis toward batch-level action-distribution regularization while retaining 1-NFE inference. Across the evaluated benchmarks and real-world tasks, it achieves the highest reported average success rates among the compared methods. The displayed ablations favor sigmoid scheduling at matched demonstration budgets. Conditional diversity, independent checkpoint evaluation, and repeated demonstration-budget trials remain directions for further validation. Matched-observation action samples could directly test conditional diversity beyond task success.

\bibliographystyle{IEEEtran}
\bibliography{references}
\par\pagebreak
\end{document}